%% file: root.tex
\documentclass[letterpaper, 10 pt, conference]{ieeeconf}  % Comment this line out if you need a4paper

\IEEEoverridecommandlockouts                              % This command is only needed if 
\usepackage{graphicx}      %
\usepackage{amsmath,amssymb}
\usepackage{wrapfig}
\usepackage{float}
\usepackage{tikz}
\usepackage{pgfplots}
\usepackage{svg}
\usepackage{ragged2e}
\usepackage[utf8]{inputenc}
\DeclareUnicodeCharacter{2212}{−}
\usepgfplotslibrary{groupplots,dateplot,fillbetween}
\usetikzlibrary{patterns,shapes.arrows}
\pgfplotsset{compat=newest}
\usepackage{nicefrac}
\usepackage{comment}
\usepackage{color}
\usepackage{caption}
\usepackage{subcaption}
\usepackage{float}
\usepackage{algorithm}
\usepackage{algpseudocode}

\algnewcommand{\LeftComment}[1]{\Statex \(\triangleright\) #1}

\title{\LARGE \bf
Optimizing Multi-Touch Textile and Tactile Skin Sensing \\Through Circuit Parameter Estimation
}

\author{Bo Ying, Su$^{1}$, Yuchen Wu$^{1}$, Chengtao Wen$^{2}$, Changliu Liu$^{1}$% <-this % stops a space
\thanks{$^{1}$Carnegie Mellon University, Pittsburgh, PA. Contact: {\tt\small \{boyings, ywu7, cliu6\}@andrew.cmu.edu}}%
\thanks{$^{2}$Siemens Corporate Technology, Berkeley, CA, 94704, USA}%
}

\begin{document}

\maketitle
\thispagestyle{empty}
\pagestyle{empty}

%%%%%%%%%%%%%%%%%%%%%%%%%%%%%%%%%%%%%%%%%%%%%%%%%%%%%%%%%%%%%%%%%%%%%%%%%%%%%%%%
\begin{abstract}

        Tactile and textile skin technologies have become increasingly important for enhancing human-robot interaction and allowing robots to adapt to different environments. Despite notable advancements, there are ongoing challenges in skin signal processing, particularly in achieving both accuracy and speed in dynamic touch sensing. This paper introduces a new framework that poses the touch sensing problem as an estimation problem of resistive sensory arrays. Utilizing a Regularized Least Squares objective function—which estimates the resistance distribution of the skin—we enhance the touch sensing accuracy and mitigate the ghosting effects, where false or misleading touches may be registered. Furthermore, our study presents a streamlined skin design that simplifies manufacturing processes without sacrificing performance. Experimental outcomes substantiate the effectiveness of our method, showing 26.9\% improvement in multi-touch force-sensing accuracy for the tactile skin.

\end{abstract}
%%%%%%%%%%%%%%%%%%%%%%%%%%%%%%%%%%%%%%%%%%%%%%%%%%%%%%%%%%%%%%%%%%%%%%%%%%%%%%%%
\begin{figure}[h]
        \centering
        \includegraphics[width=\columnwidth]{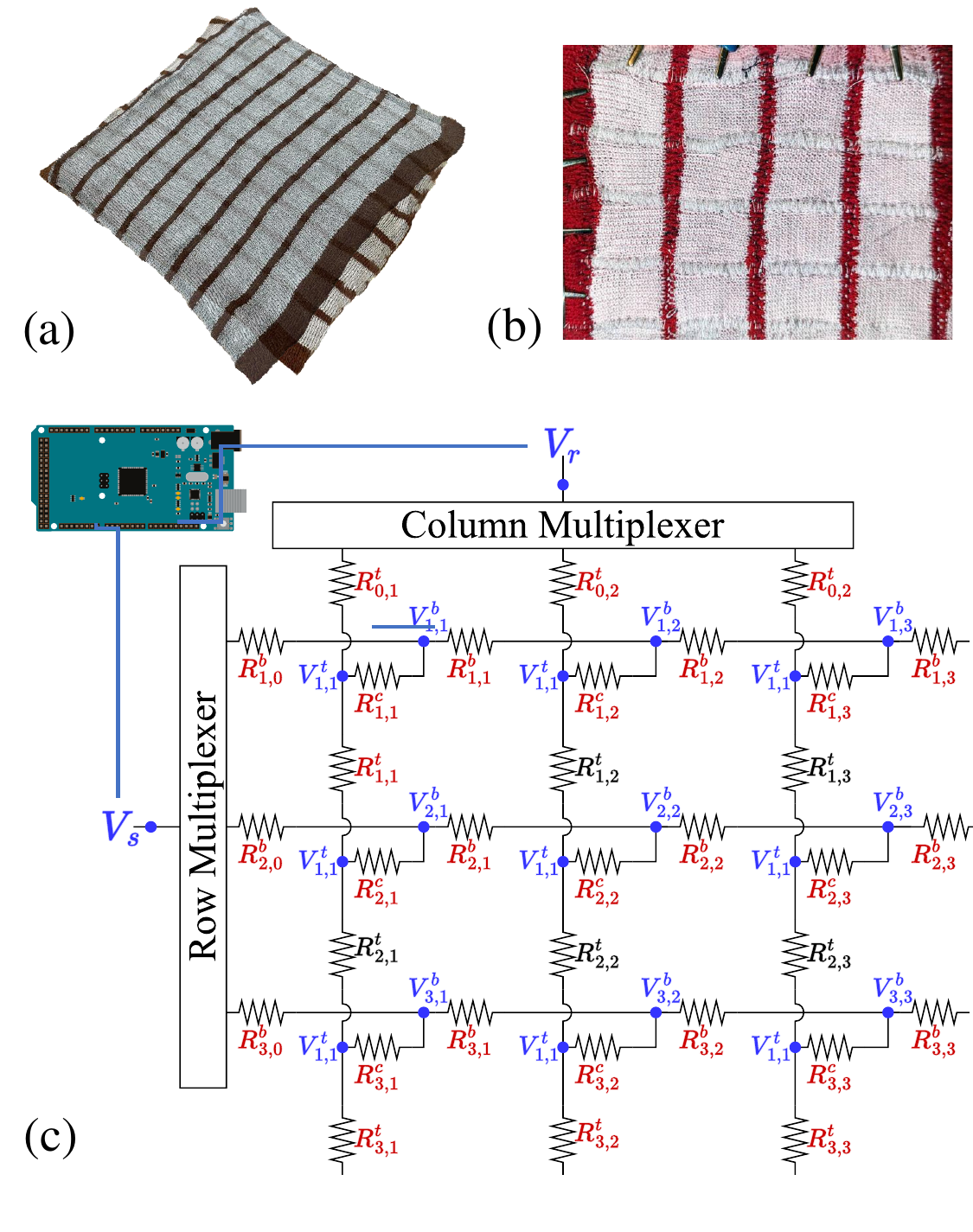}
        \vspace{-25pt}
        \caption{ An overview of our proposed tactile skin sensing method.  (a) Two textile pieces with conductive stripes are separately knitted.
        (b) The two textile pieces are sewn together orthogonally to create a grid of sensing cells. The light pink fabric is made of Nylon Stretchy Yarn and the red vertical fabric stripes are made of Acrylic Yarn. (c) The skin is modeled as a resistive sensory array. Our approach predicts force applied on the skin by estimating cell resistances $R^{C}$ using the Arduino board.}
        
        \label{fig: paper-overview}
\end{figure}
\section{INTRODUCTION}
% \subsection{Motivation}
% We need an advanced signal processing framework to improve the accuracy of tactile skin and compensate for the undesired effects due to the physical properties of the skin.
%\section{INTRODUCTION}

As robotics become increasingly integrated into everyday life, tactile and textile skins have emerged as essential technologies for ensuring safe and nuanced human-robot interactions \cite{Lee_2021}. While textile-based skins offer benefits like scalability, low cost, and adaptability \cite{robotsweater}, they present signal processing challenges, including nonlinearity and signal ghosting, that hinder their practical application.

This paper introduces a new signal-processing framework designed to improve the \textit{accuracy} and \textit{reliability} of the multi-touch force sensing of textile-based tactile skins. We proposed the very first solution to achieve accurate multi-touch textile-based tactile sensing relying entirely on textile materials. We formulate the textile-based tactile sensing problem as a parameter estimation problem for a resistive sensor array, which can be solved via optimization techniques. This work focuses on the resistive sensing approach on textiles due to its simplicity and robustness against deformation. Our framework requires only one single-point force calibration in the beginning and can make accurate multi-touch force predictions afterward. Our approach also effectively mitigates \textit{ghosting} by assuming that any sensing cell can influence another, irrespective of the touch patterns. 

Due to the enhanced multi-touch sensing capability, we further simplified the tactile skin to a two-layer structure. The two-layer skin is subject to a more serious ghosting effect due to direct contact between two conductive layers despite its advantage of a much lower minimum detectable force. As our solution considers a more general multi-touch scenario, our signal-processing framework has significantly mitigated the undesired ghosting effect for the two-layer skin, and therefore, unleashed its potential for lower minimum detectable force. Experiments later demonstrated the lower detectable force based on our integrated hardware-software solution.

The remainder of this paper is organized as follows: Section II reviews related work in tactile skins and existing resistance estimation methods; Section III describes our novel design for 2-layer textile-based tactile skin; Section IV outlines our problem statement and proposed methodology; Section V presents and discusses our experimental results; Section VI concludes the paper.

% \subsection{Contributions}
% We formulate the problem as a circuit component parameter estimation problem. We solve an optimization problem with a Regularized least squares objective function for improved force sensing accuracy. This allows accurate readings even if the skin is varying (can be stretched, bent, etc.)
% Improved scanning method & reference resistors increasing dynamic range
% Simplified skin design (2 layers) for easier manufacturing
% \subsection{Organization of the Paper}
% We first discuss the background and related work in Section II. We then discuss the design and properties of the skin in Section III. We then formulate the problem in Section IV. We then present our proposed method in Section V. We then present the experimental evaluation in Section VI. We then discuss the results in Section VII. We then conclude the paper in Section VIII.

%%%%%%%%%%%%%%%%%%%%%%%%%%%%%%%%%%%%%%%%%%%%%%%%%%%%%%%%%%%%%%%%%%%%%%%%%%%%%%%%

\section{BACKGROUND AND RELATED WORK}
\subsection{Textile and Tactile Skin}
% Functions: can sense pressure
% Applications: robot-human interaction, safety control, robot control
% Properties: stretchable, flexible, robust
% Fabrication: low cost, easy to manufacture
Tactile skins provide robots with the capability to sense pressure, facilitating a wide range of applications from nuanced human-robot interaction to advanced safe control mechanisms \cite{Cirillo2016, Liang2020}. Various fabrication techniques of tactile skins have been proposed, including capacitive units featuring flexible polyethylene terephthalate (PET) substrates \cite{Ji2001}, and piezoresistive arrays with star-shaped strain gauges \cite{Saadatzi2019}. A detailed review of different manufacturing techniques and effective utilization of tactile skins can be found in this survey \cite{6583342}.

Recent advancements in textile-based tactile skins offer benefits like scalability, cost-effectiveness, and stretchability \cite{robotsweater, su2023customizing}. However, they introduce unique challenges in signal processing. One significant issue is the non-negligible wire resistances in conductive textiles, which makes the accurate estimation of contact forces more complex. Another distinct challenge arises from the alternate paths of current, which can lead to false touches or ghosting when multiple simultaneous contacts occur.

This paper aims to address these specific challenges through a novel signal-processing framework, thereby improving the performance and reliability of textile-based tactile skins.

\subsection{Readout of Resistance Sensor Arrays}
% Skin itself:
% Predicts force based on the raw voltage drop due to the change in resistance
% Limitations: ghosting effects, resolution, cannot adapte to the dynamic nature of the skin, noise, which would be described in the next section

% Resistor network:
% Hardware-based approach
% Limitations: More complex, more expensive, more difficult to manufacture
% defeats the purpose of using tactile "skin" (stretchable, flexible ...etc)
% Software-based approach
% Limitations: cannot adapt to the dynamic nature of the skin, does not consider wire resistances
% Requires a lot of calibration data, does not work well with dynamic skin

% The force estimation of the textile-based tactile skin originally relies on measuring the voltage drop caused by change of resistance between skin layers. However, voltage drop at the reference resistors could be caused by multiple factors 
Our approach to estimating multiple contact forces relies on accurately determining the resistance in each cell of a tactile skin array, connected in rows and columns as shown in Figure \ref{fig: paper-overview}(c). The challenge of accurately reading resistance sensor arrays has been extensively studied along with many different approaches~\cite{SAXENA2009,Wu2017,ZHANG2023}.

Existing solutions to the readout problem generally fall into two categories. The first involves the addition of extra electrical components to the array for better resistance estimation. For example, Snyder et al inserted diodes to calculate the current in each column's sensitive elements and, subsequently, their resistance~\cite{Snyder1978}. Tanaka et al introduced transistors into a 128×128 resistive array for the same purpose~\cite{Tanaka}. While effective, these hardware modifications introduce complexities, inflate costs, and compromise the skin's flexibility and stretchability.

The second approach employs a resistance matrix, first suggested in \cite{Shu2015} and later refined in \cite{IRMA}, to estimate individual resistances. While this approach is less complex than hardware modifications, it fails to account for wire resistances and the uneven resistance distribution of textile-based tactile skins.

In light of these limitations, we propose the first optimization-based approach for textile-based tactile skins. Unlike conventional approaches, our method does not assume constant wire resistances, which can change due to stretching or bending. As a result, our approach enhances the accuracy of contact force estimation, a claim substantiated by our comprehensive experimental results.

% \subsection{Literature Gap}
% No existing work takes the optimization approach to improve the accuracy of tactile skin.

%%%%%%%%%%%%%%%%%%%%%%%%%%%%%%%%%%%%%%%%%%%%%%%%%%%%%%%%%%%%%%%%%%%%%%%%%%%%%%%%
\section{TEXTILE AND TACTILE SKIN: \\DESIGN AND PROPERTIES}
Our design builds upon the tactile skin model proposed by Si. et al \cite{robotsweater}, but incorporates several modifications. Unlike the three-layered construction suggested in their work, we employ a \textit{simplified two-layer} construction by removing the insulating layer. Our two-layer construction was chosen for its ease of manufacturing and lower minimum detectable force. The readout frequency of the tactile skin depends on the grid dimensions of the skin, where 8x8 skin runs at a 90Hz readout rate.

\subsection{Working Principle}
% The contact between the top and bottom layer of conductive fabric causes a change in resistance, which causes a change in voltage drop across the resistor network, which and be detected by the ohmmeter circuit.
The functionality of our tactile skin is dependent on the interaction between its two layers of conductive fabric. Upon applying pressure, these layers come into contact, inducing a change in resistance. This change influences the voltage across an associated resistor network, which can be measured through a resistance measurement circuit. As more pressure is exerted, the contact area between the layers expands, reducing the resistance and consequently altering the voltage drop across the resistor network. This voltage drop serves as an estimator for the applied force.
% Figure showing pressure vs voltage line plot
\subsection{Skin Material and Fabrication}

% Material: Non-conductive fabric and conductive fabric
Our tactile skin is made from three different types of fabrics, each serving a specific purpose:  Acrylic Yarn, the red portion shown in Figure \ref{fig: paper-overview} (b) offers structure and durability. Nylon Stretch Yarn (MaxiLock Stretch Textured Nylon), adds stretchability, and Bekinox-polyester Stainless Steel Yarn (Baekert BK 9036129) provides conductivity. The Nylon Stretch Yarn and Stainless Steel Yarn are co-knitted to form the conductive stripes of the textile skin, which can be seen in the light color portion in Figure \ref{fig: paper-overview} (b). By executing the knitting program written in the knitout language \cite{knitout}, the skin's top and bottom layers are first knitted individually with stripe patterns. They are then joined along the edges of the conductive stripes using a sewing machine, as illustrated in Figure \ref{fig: paper-overview}.
% Fabrication: Knitting machine, sewing machine
\subsection{Properties and Characteristics}
% Stretchability numbers in %, vertical and horizontal
% resistance-per-unit-length vertical and horizontal in ohms/m
% Figure: Stretch % vs resistance-per-unit-length scatter plot
% Figure: Voltage vs pressure scatter plot
The skin's minimum detectable pressure is primarily determined by two factors: the inherent resistance of the conductive yarn, which is specified as \(20 \, \Omega/\text{cm}\) for Baekert BK 9036129, and the deformation of the skin under applied pressure. The stretchable properties of the yarn, combined with the selected knitting pattern, confer bidirectional stretchability to the skin. It should be noted that the resistance within the skin exhibits spatial variations. Specifically, the resistance per unit length tends to be lower along the yarn compared to that between stitches, resulting in a non-uniform distribution of resistance. The resistance is generally lower in the course (horizontal) direction compared to the wale (vertical) direction \cite{uknit}. Furthermore, the resistance values lower as the skin stretches, adding complexity to the skin's resistance profile \cite{uknit}. In this work, it is not required to measure actual resistances of the skin as it is solved numerically.

\subsection{Challenges in Tactile Sensing}
Multi-touch tactile sensing in textile-based sensors presents multiple challenges, including undesired ghosting effects and fluctuations in skin resistance. Moreover, to preserve key attributes like manufacturability, stretchability, and flexibility, it's crucial to avoid adding extraneous components. Maintaining a fabric-only composition not only ensures the washability of the tactile skin but also minimizes the risk of electronic damage. Given these design considerations, the following challenges are intrinsic to the development of textile-based tactile sensors.

\subsubsection{Ghosting Effects}
\begin{figure}[t]
         \centering
     \vspace{10pt}
     \begin{subfigure}[t]{0.60\linewidth}
         \centering
                \includegraphics[width=\columnwidth]{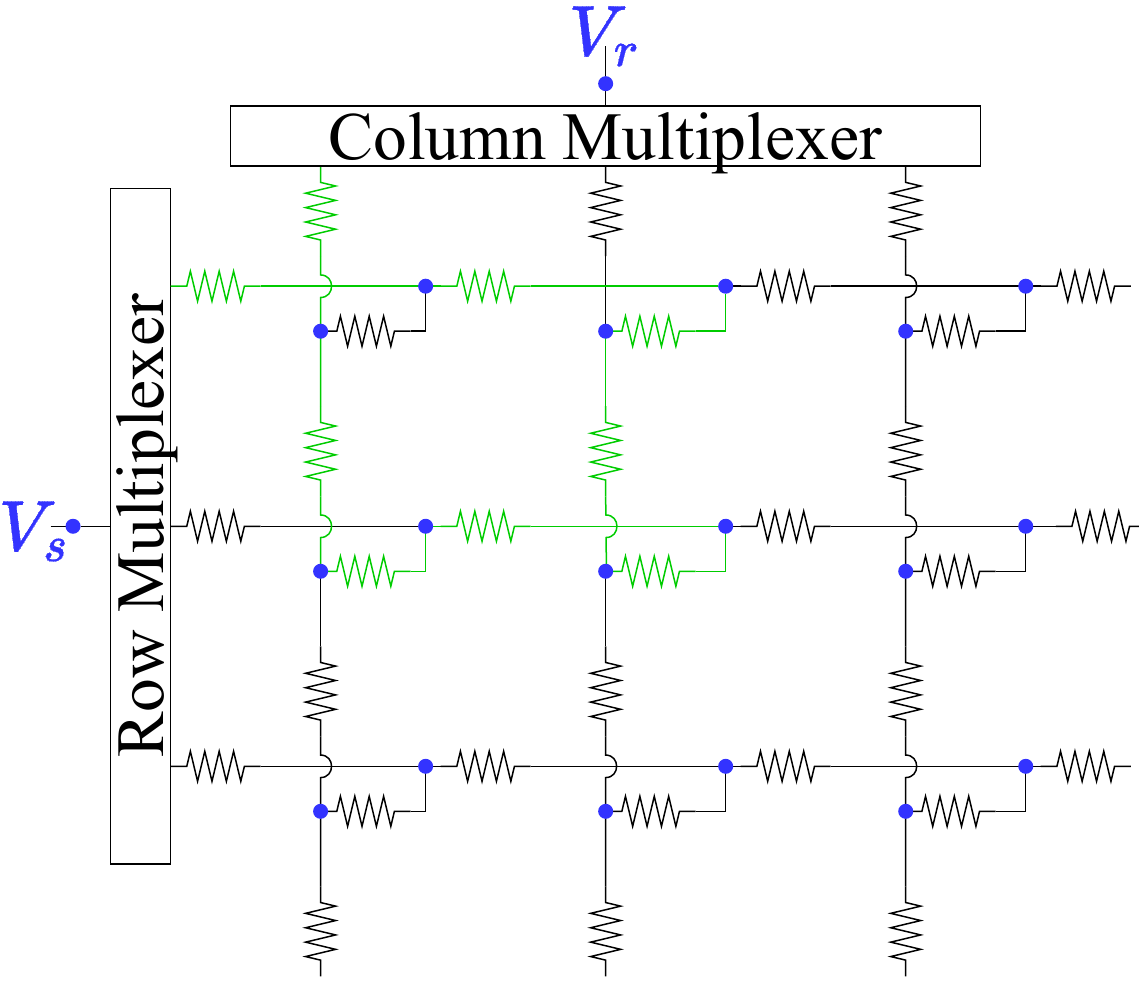}
                % \label{fig:ghosting-illustration}
                \caption{}
     \end{subfigure}
     \hfill
     \begin{subfigure}[t]{0.34\linewidth}
         \centering
        \includegraphics[width=\columnwidth]{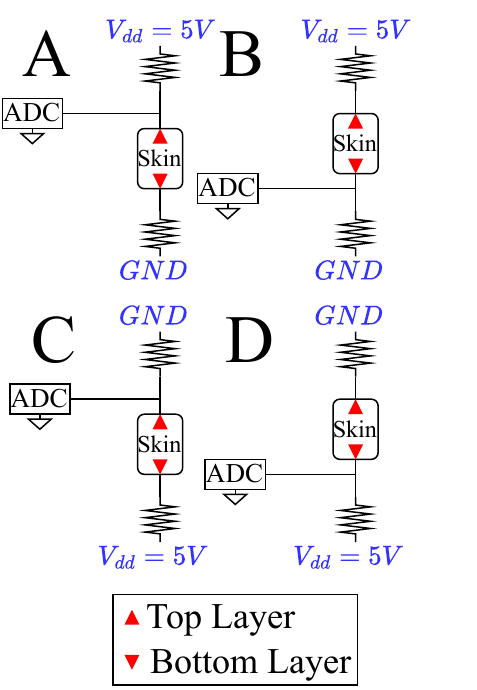}
        \caption{}
        % \label{fig: voltage-divider}
     \end{subfigure}
     \vspace{10pt}
     \caption{(a): Ghosting effects occur when an alternate path of current (shown in green) is formed for the sensing cell at the upper left corner, bypassing it completely. (b): Ohmmeter Configurations. Details are provided in Section IV(C).}
    \label{fig:ghosting-voltage}
\end{figure}

As shown in Figure \ref{fig:ghosting-voltage} (a), false positive detections occur due to alternate electrical paths between the skin layers. When three or more touches align on a rectangle's corners, the untouched corner mistakenly registers a touch, a result of currents flowing an alternate path between the layers.

\subsubsection{Adaptation to varying skin resistance distribution}
The tactile skin's stretchability and flexibility introduce resistance variations along the conductive stripes, which skews the skin output in multi-touch scenarios. Traditional resistor matrix methods, designed to combat ghosting, operate under assumptions that may not apply to textile and tactile skins. Changes in resistance, due to stretching or bending, challenge accurate force estimation.
%%%%%%%%%%%%%%%%%%%%%%%%%%%%%%%%%%%%%%%%%%%%%%%%%%%%%%%%%%%%%%%%%%%%%%%%%%%%%%%%
% \begin{figure}[t]
%         \begin{center}
%                 \centering
%                 \includegraphics[width=\columnwidth]{Figures/skin_circuit_model.pdf}
%                 \caption{The Circuit Model for the tactile skin. We assume that the readout voltage $V_s$ and $V_r$ could be fully explained by cell and wire resistances from the top and bottom layers} 
%                 \vspace{-1\baselineskip}
%                 \label{fig: skin-circuit-model}
%         \end{center}
% \end{figure}

% \begin{figure}[t]
%         \begin{center}
%                 \centering
%                 \includegraphics[width=\columnwidth]{Figures/dRoverR.png}
%                 \caption{Resistance measurement error over ADC range}
%                 \vspace{-1\baselineskip}
%                 \label{fig: dRoverR}
%         \end{center}
% \end{figure}

\begin{algorithm}[b]
\caption{Resistance Estimation}
\begin{algorithmic}[1] % The number tells where the line numbering starts
\Procedure{Resistance Estimation}{}
   \LeftComment{Step 1: Take voltage measurement}
   \State $V_s, V_r \gets \text{four measurements for each cell}$
   \LeftComment{Step 2: Calculate initial circuit states}
   \State $R^t, R^b \gets 0 $
   \State $R^c \gets \text{results solved with Ohm's law} $
   \State $V^t, V^b \gets Simulation(V_s, V_r,R^t, R^b, R^c)$
   \State $P \gets \{V_s, V_r,R^t, R^b, R^c\}$
   \LeftComment{Step 3: Solving WithOUT Regularization}
   \State $P\gets solveLeastSquares(P)$
   \LeftComment{Step 4: Solving With Regularization}
   \State $P\gets solveRegularized(P)$
\EndProcedure
\end{algorithmic}
\label{resistance_estimation}
\end{algorithm}

\section{Algorithms for Multi-Touch Sensing}
% Figure: Circuit diagram
% Figure: Ohmmeter circuit
% We simplify the physical structure of the skin into a circuit model that consists of resistors and voltage sources. The circuit model reflects the physical construction of the skin, where the top and bottom layers of the skin are consists of conductive stripes. The conductive stripes are modeled as resistors with unknown resistances connected in series, and the contact between the stripes on the top and bottom layers is modeled as a grid of resistors with unknown resistances.
% The measurement is done using a ohmmeter circuit, multiplexed across rows and columns. The voltage source is connected to reference resistor to one of the input to the skin resistor network in series, and then connected to another reference resistor before reaching the ground. The symmetry of the circuit allows us to use either the top or bottom layer as source or ground.
% We formulate the problem as a circuit component parameter estimation problem, where the unknown resistances are the target parameters. By applying different voltage to the top and bottom layers, we obtain the total voltage drop across the entire skin resistor network. We then solve an optimization problem to estimate the unknown resistances.
% The resistances can then be used to calculate the force applied to the skin.
\subsection{Modeling}
We model the physical tactile skin structure as a simplified circuit comprising resistors and voltage sources shown in Figure \ref{fig: paper-overview}(c). This circuit model mirrors the tactile skin's construction, where both the top and bottom layers consist of conductive stripes. These conductive stripes are represented as resistors connected in series with unknown resistances, $R_{i,j}^t$ for the top layer and $R_{i,j}^b$ for the bottom layer where $i$ is the row index and $j$ is the column index. The point of contact between the top and bottom stripes is similarly modeled as a grid of resistors with unknown resistances $R_{i,j}^c$, whose value changes when different contact forces are applied. A voltage source $V_{dd}$ is connected through a reference resistor, multiplexed to one input to the skin resistor network, and is then linked to another reference resistor before reaching the ground. The problem of contact/force estimation problem is formulated as a parameter estimation problem, where we aim to solve unknown sensing cell resistances $R_{i,j}^{C}$. By applying different voltages to the top and bottom layers, we obtain the total voltage drop across the skin's resistor network from pairs of electrodes across rows and columns. We then employ optimization techniques to estimate these unknown resistances, which are subsequently used to predict the force applied to the skin.

%%%%%%%%%%%%%%%%%%%%%%%%%%%%%%%%%%%%%%%%%%%%%%%%%%%%%%%%%%%%%%%%%%%%%%%%%%%%%%%%

%\section{METHODS}
\subsection{Overview of the Solution Approach}
% Figure: Overview of the proposed method, system flow diagram
% System flow diagram:
% In each cycle, we take four measurements with different ohmmeter configurations for each cell.
% Assuming the wire resistances are 0, we calculate the cell resistances using the ohmmeter formula for each cell.
% For each measurement, we run circuit simulations with guessed values for the unknown resistances to get the voltage readings at each nodes in the circuit model.
% We then create an optimization program that models the circuit and the measurement process. The guessed resistances and simulated voltage readings are used as the starting point for the optimization program.
% We first solve for feasibility, then we solve for optimality.
% The solved resistances are then used to calculate the force.

Our approach to multi-touch force sensing using textile-based tactile skin comprises two stages. The first stage performs one-time single-touch force calibration. Given the readout data acquired from the Ohmmeter circuit, we estimate the cell resistances $R_{i,j}^c$ of the skin using the proposed resistance estimation algorithm shown in Algorithm \ref{resistance_estimation} and fit them to simple linear regression models for each cell, which predicts the applied force. The second stage is the estimation stage, where the skin cell resistances are solved online using the same optimization program. Multi-touch forces are predicted by the regression model created in the first stage using the solved resistances.

\subsection{Readout Process}
% For each cell, we take 4 measurements with different ohmmeter configurations.
% Figure: 4 different ohmmeter configurations
% Figure: 4 measurements vs multi-meter ground truth

In the readout process, we first take four distinct measurements for each cell using different ohmmeter configurations. The four distinct ohmmeter configurations, labeled A, B, C, and D, are employed for each sensing cell as shown in Figure 2(b). This dual-layer strategy effectively compensates for reference resistor tolerances and addresses the non-ideal behavior of tactile skin by eliminating systematic errors.

\subsection{Resistance Estimation}
In both calibration and estimation stages, we are using the same procedure to solve for unknown sensing cell resistances $R_{i,j}^C$, as shown in Algorithm \ref{resistance_estimation}. 
The process of resistance estimation is as follows:
Given the measurements for each cell using different ohmmeter configurations in the readout process, we create circuit state variables by applying the circuit model of the skin shown in Figure \ref{fig: paper-overview}(c), aiming to find a set of unknown resistance and voltage values  \( R_{ij}^T \), \( R_{ij}^B \), \( R_{ij}^C \), \( V_{ij}^T \), and \( V_{ij}^B \) that best explains the observation. 
We bootstrap the optimization program by calculating initial resistances for each cell assuming zero wire resistances. 
Following this, we run circuit simulations using the estimated resistances to obtain the initial voltage at each circuit node. 
These initial values provide a starting point for an optimization program designed to refine the estimated resistances and voltages. 
The optimization process consists of two stages: the first focuses on finding a least squares solution that complies with Kirchhoff's Law and Ohm's Law shown in step 3, while the second stage aims to regularize wire resistances to yield regularized values shown in step 4.

\subsection{Proposed Optimization Program}

We define the decision variables to the proposed optimization program as a set of circuit states, each of which corresponds to one measurement using the Ohmmeter circuit. A circuit state of each measurement consists of 5 sets of variables: \( R_{ij}^t \), \( R_{ij}^b \), \( R_{ij}^c \), \( V_{ij}^t \), and \( V_{ij}^b \).

Our optimization program enforces circuit equality constraints for each circuit state based on Kirchhoff's Current Law and Ohm's Law. This ensures any feasible solution the solver returns to also be a valid solution to the circuit model.

\subsection{Least Squares Objective Function}
We design the objective function to be the sum of the squared discrepancies between the resistance values among the circuit states. 
We define two least squares objectives.
The first objective $Cost_{f}$ is the sum of discrepancies between the $A$ and $B$ configuration and the sum of discrepancies between the $C$ and $D$ configuration. Let
\begin{equation}
        ^{k}P=\{^{k}R^t\cup\ ^{k}R^b\cup
        \ ^{k}R^c\cup\ ^{k}V^t\cup\ ^{k}V^b\}
\end{equation}
be the set of unknown variables in the circuit states with ohmmeter configuration $k$. We define
\begin{equation}
        Cost_{f} = \sum_{p\in P} (^{A}p\ 
        -\  ^{B}p)^2 + (^{C}p\ -\ ^{D}p)^2
\end{equation}
Since $A$ and $B$ configurations and $C$ and $D$ configurations consist of consecutive measurements with the same input voltage configuration, we assume both the voltages and resistance values would not change much between the two measurements.

The second objective $Cost_{c}$ is the sum of discrepancies between the circuit states over measurements in all the cells with the same ohmmeter configurations. Let
\begin{equation}
        ^{ij}Q=\{^{ij}R^t\cup\ ^{ij}R^b\cup\ ^{ij}R^c\}
\end{equation}
be the set of unknown resistance variables in the circuit states when measuring the cell at column $i$ and row $j$. We define
\begin{equation}
        Cost_{c} = \sum_{i=1}^{n} \sum_{j=1}^{m} \sum_{q\in Q} (^{ij}q\ -\ ^{i(j+1)}q)^2
\end{equation}
Note that the comparison is conducted in column-major order, which aligns with the readout sequence. For example, when solving a $2\times2$ skin, a squared cost is defined for the discrepancy between the circuit state when measuring the cell at (1, 1) using ohmmeter configuration \(A\) and the circuit state measuring (2, 1) under the same ohmmeter configuration \(A\). This approach is based on our assumption that the resistances remain stable between consecutive measurements.
The full least squares objective function is defined as follows:
\begin{equation}
        Cost_{lsq} = \alpha Cost_{f} + \beta Cost_{c}
\end{equation}
where $\alpha$ and $\beta$ are hyper-parameters where we set both to 1.0e6.

\subsection{Regularization}
% Figure showing the effect of regularization (line plots)
% Explain the idea of penalizing large wire resistances
Since the circuit states create underdetermined systems of equations, we add a regularization term to the objective function to ensure the solution is unique. \
We applied the $L_2$ regularization term to the objective function that penalizes large wire resistances in the top and bottom stripes. For each circuit state, a regularization term is added, defined as
\begin{equation}
        Cost_{r} = \sum_{i=1}^{n} \sum_{j=1}^{m} (R_{ij}^t)^2 + (R_{ij}^b)^2
\end{equation}
As a result, the optimization program ensures both the circuit is feasible by complying with Kirchhoff's current law, and the wire resistances are small for each circuit state, as the force signal would be more likely to be attributed to changing cell resistances. The regularized objective function is defined as
\begin{equation}
        Cost_{reg} = \alpha Cost_{f} + \beta Cost_{c} + \lambda Cost_{r}
\end{equation}
where $\lambda$ is the regularization weight where we set it to 1.0e9. We also set $\alpha = 1$, $\beta = 1$. In practice, we found that as long as the skin itself is not stretched or bent too violently in a short time, the solver can find local optimum solutions quickly using the previous solution as a starting point.

\subsection{Optimization}
% We use the interior-point algorithm to solve the optimization problem
% We provide feasible initial values for the decision variables by running simulations on the circuit model with guessed values for the unknown resistances.
% We solve for feasibility first, then we solve for optimality.
In our optimization process, we employ the interior-point algorithm to find the best-fitting circuit parameters using the Ipopt solver \cite{wachter2006implementation}. Given the initial resistances obtained using Ohm's Law and initial voltages obtained using a circuit simulator~\cite{hayes2022lcapy}, we first solve the optimization program without the regularization term for least squares objective $Cost_{lsq}$, then solve the program with the regularization objective $Cost_{reg}$.
We scale the resistance values close to 1 to maximize machine floating-point precision and use the inverse of resistance to represent constraints in the optimization problem.

\subsection{Force Prediction}
In the calibration stage, we collect single-point calibration data using a robotic arm with a Robotiq FT 300-S Force Torque Sensor attached to its end effector, which the force sensor has a signal noise of 0.1 N. We create linear regression models for each sensing cell using the calibration data. We rely on the single-point calibration models to predict force given the solved resistance values in the estimation stage, where multiple touches may present.

%%%%%%%%%%%%%%%%%%%%%%%%%%%%%%%%%%%%%%%%%%%%%%%%%%%%%%%%%%%%%%%%%%%%%%%%%%%%%%%%
\begin{figure}[t!]
     \vspace{10pt}
        \centering
        \begin{tabular}{cc}
                
                \rotatebox{90}{\parbox{1cm}{\small Ground \\ Truth}}
                \includegraphics[width=0.15\linewidth]{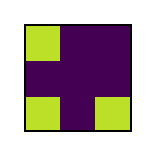}
                \includegraphics[width=0.15\linewidth]{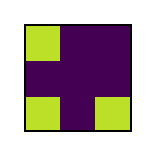}
                \includegraphics[width=0.15\linewidth]{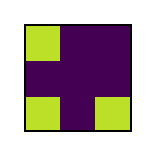}
                \includegraphics[width=0.15\linewidth]{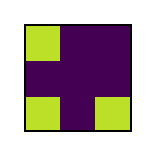}
                \includegraphics[width=0.15\linewidth]{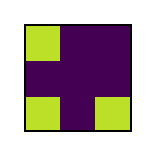}\\
                \rotatebox{90}{\parbox{1cm}{\small Naive \\ Solution}}              
                \includegraphics[width=0.15\linewidth]{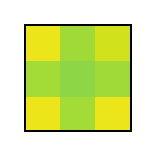}
                \includegraphics[width=0.15\linewidth]{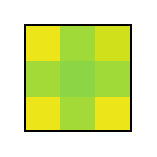}
                \includegraphics[width=0.15\linewidth]{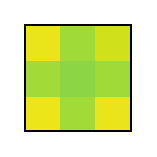}
                \includegraphics[width=0.15\linewidth]{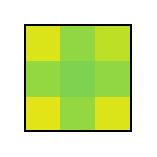}
                \includegraphics[width=0.15\linewidth]{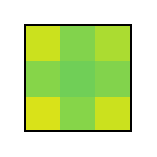}\\
                \rotatebox{90}{\parbox{1cm}{\small Feasible \\ Solution}}   
                \includegraphics[width=0.15\linewidth]{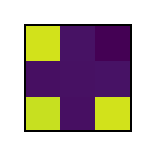}
                \includegraphics[width=0.15\linewidth]{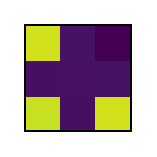}
                \includegraphics[width=0.15\linewidth]{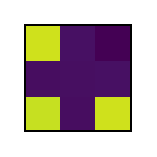}
                \includegraphics[width=0.15\linewidth]{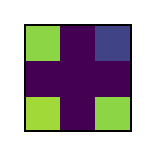}
                \includegraphics[width=0.15\linewidth]{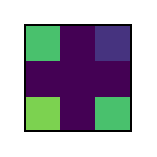}\\
                \rotatebox{90}{\parbox{1cm}{\small Optimal \\ Solution}}   
                \includegraphics[width=0.15\linewidth]{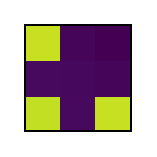}
                \includegraphics[width=0.15\linewidth]{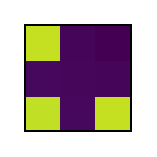}
                \includegraphics[width=0.15\linewidth]{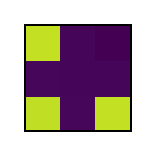}
                \includegraphics[width=0.15\linewidth]{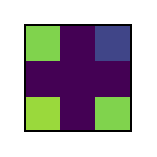}
                \includegraphics[width=0.15\linewidth]{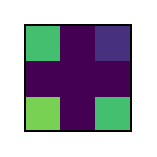}\\
                \resizebox{0.9\linewidth}{!}{\input{Figures/simulation_wiresweep.tex}}\\
        \end{tabular}
\caption{Simulation results for varying wire resistances in tactile skin. Brighter colors denote lower resistances, ranging from $0 \Omega$ to $1.0 M\Omega$. The pressed cells are assigned a lower\(0.001 M\Omega\) resistance, while unpressed cells have higher\(1.0 M\Omega\). Wire resistances vary as \(0.0001-0.041 M\Omega\) from left to right. These values are typical for our setup.}

        \label{fig:simulation-wiresweep}
        
\end{figure}%

\begin{figure}[t!]
     \vspace{10pt}
        \centering
        \begin{tabular}{cc}
                \rotatebox{90}{\parbox{1cm}{\small Ground \\ Truth}}
                \includegraphics[width=0.15\linewidth]{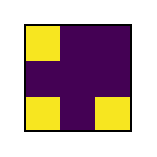}
                \includegraphics[width=0.15\linewidth]{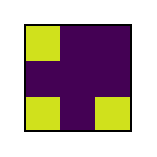}
                \includegraphics[width=0.15\linewidth]{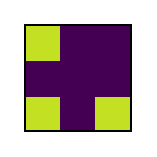}
                \includegraphics[width=0.15\linewidth]{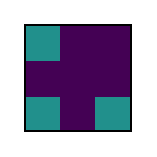}
                \includegraphics[width=0.15\linewidth]{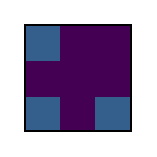}\\

                \rotatebox{90}{\parbox{1cm}{\small Naive \\ Solution}}
                \includegraphics[width=0.15\linewidth]{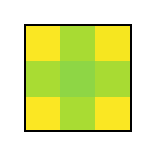}
                \includegraphics[width=0.15\linewidth]{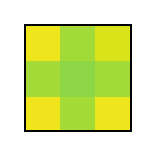}
                \includegraphics[width=0.15\linewidth]{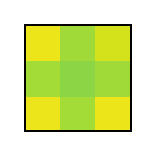}
                \includegraphics[width=0.15\linewidth]{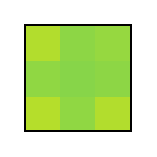}
                \includegraphics[width=0.15\linewidth]{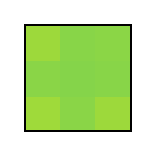}\\
                \rotatebox{90}{\parbox{1cm}{\small Feasible \\ Solution}}
                \includegraphics[width=0.15\linewidth]{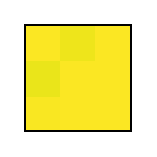}
                \includegraphics[width=0.15\linewidth]{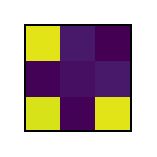}
                \includegraphics[width=0.15\linewidth]{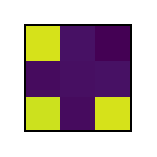}
                \includegraphics[width=0.15\linewidth]{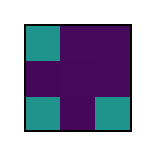}
                \includegraphics[width=0.15\linewidth]{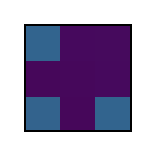}\\
                \rotatebox{90}{\parbox{1cm}{\small Optimal \\ Solution}}
                \includegraphics[width=0.15\linewidth]{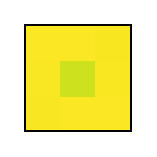}
                \includegraphics[width=0.15\linewidth]{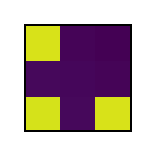}
                \includegraphics[width=0.15\linewidth]{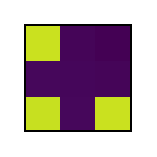}
                \includegraphics[width=0.15\linewidth]{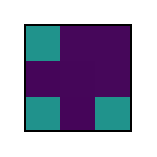}
                \includegraphics[width=0.15\linewidth]{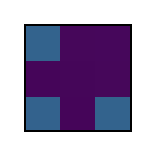}\\
                \resizebox{0.9\linewidth}{!}{\input{Figures/simulation_cellsweep}}\\
        \end{tabular}
\caption{Simulation results for tactile skin with variable cell resistances. Brighter colors denote lower resistances, ranging from $0 \Omega$ to $1.0 M\Omega$. The pressed cells are simulated with resistances of \(0.01- 0.7 M\Omega\) displayed from left to right. Unpressed cells have a fixed resistance of \(1.0 M\Omega\), and wire resistances are \(0.001 M\Omega\). These values are typical for our setup.}

        \label{fig:simulation-cellsweep}
\end{figure}%
\section{EXPERIMENTAL EVALUATION}

To validate our method, we have conducted several experiments in simulation and with real robots. 
% We first verify that our approach mitigates the ghosting effects and accurately estimates the cell resistance values in a custom simulation. We then evaluate the performance of our method by comparing the force derived by our pipeline with the ground truth force readings measured by a force torque sensor. Last, we compare our results with the existing approach to show that our method is less prone to the ghosting effects of the circuit.
% First, we validate our algorithm against the simulation result to show that our approach mitigates the ghosting effects and improves the estimate of the resistance values.
% Second, We evaluate the performance of our method by comparing the force readings with the ground truth force readings measured by a force torque sensor.
% Third, we compare our results with the existing approach to show that our approach improves the accuracy of the tactile skin.
% Finally, we discuss dynamic rang2e and the limitations of our approach.
\subsection{Simulation Validation}
We use the circuit simulation tool~\cite{hayes2022lcapy} to assess our optimization algorithm for cell resistance estimation. Using the prior circuit model, we simulate the ghosting effect by varying wire resistances between 0.0001 $M\Omega$ to 0.041 $M\Omega$ and cell resistances from 0.01 $M\Omega$ to 0.7 $M\Omega$. 

Results on the effects of wire resistance variation are in Figure \ref{fig:simulation-wiresweep}. Here, three cells with lower resistances were pressed, revealing anticipated ghosting, especially in the top-right cell as shown in Figure \ref{fig:simulation-wiresweep} Naive Solution row. In comparison, our algorithm reduces ghosting, evident in the lower Root Mean Square Error (RMSE) in Figure \ref{fig:simulation-wiresweep} Feasible Solution row. While the regularized solution in the Optimal Solution row introduces a regularization term to further reduce RMSE, it underperforms in scenarios with high wire resistances.

To gauge accuracy in varying cell resistances with constant wire resistances, results are in Figure \ref{fig:simulation-cellsweep}. The naive approach struggles with high cell resistances as depicted in Figure \ref{fig:simulation-cellsweep} Naive Solution row. Conversely, our method recovers resistances with low RMSE, shown in Figure \ref{fig:simulation-cellsweep} Feasible Solution row. The solution in the Optimal Solution row further optimizes RMSE but struggles when cell resistances are below wire resistances, aligning with prior wire resistance findings.

\begin{figure}[t!]
         \centering
    
     \begin{subfigure}[t]{0.28\linewidth}
         \centering
        \includegraphics[width=\columnwidth]{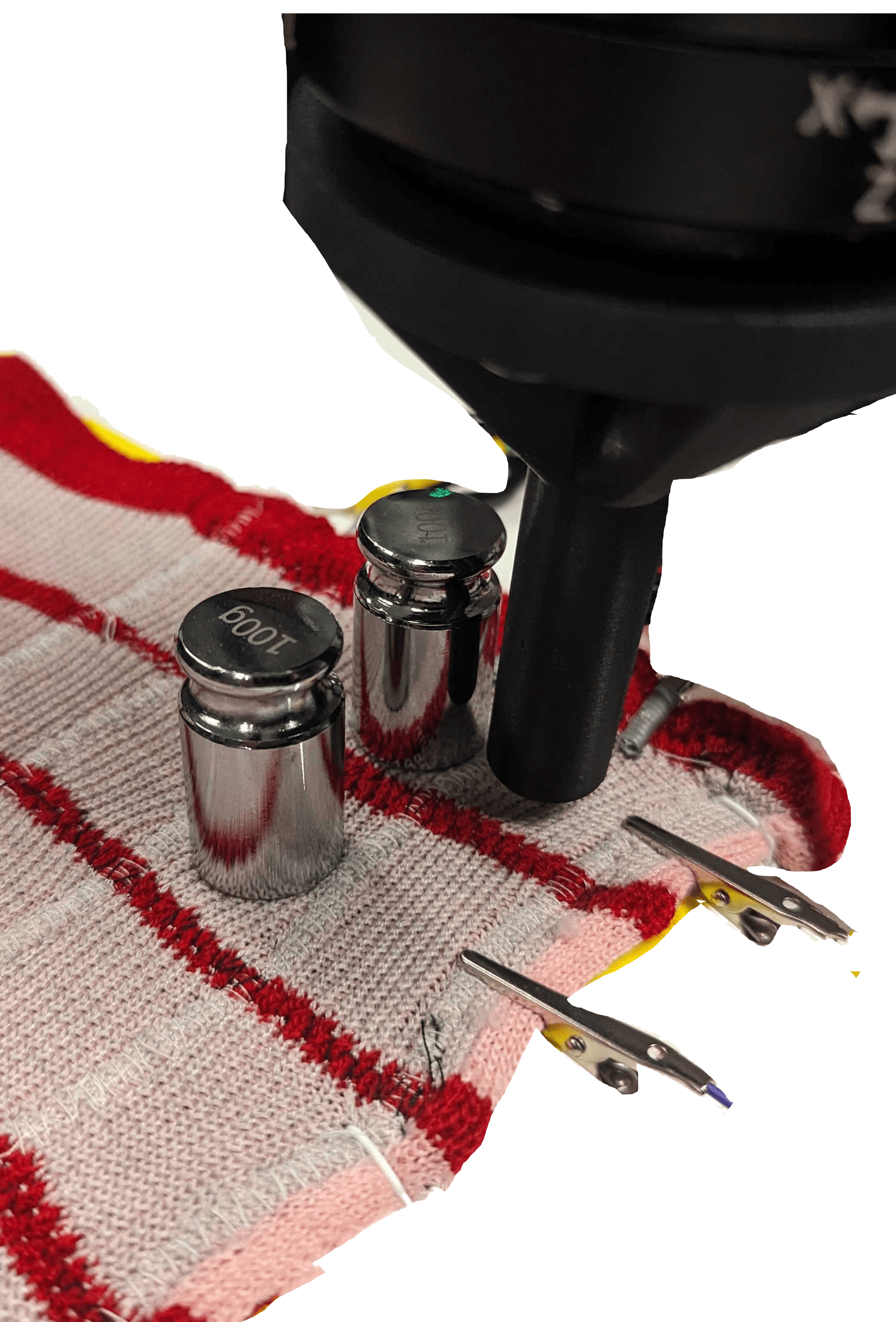}
        \caption{}
        
        \label{fig: flat-touch}
     \end{subfigure}
     \hfill
     \begin{subfigure}[t]{0.27\linewidth}
         \centering
        \includegraphics[width=\columnwidth]{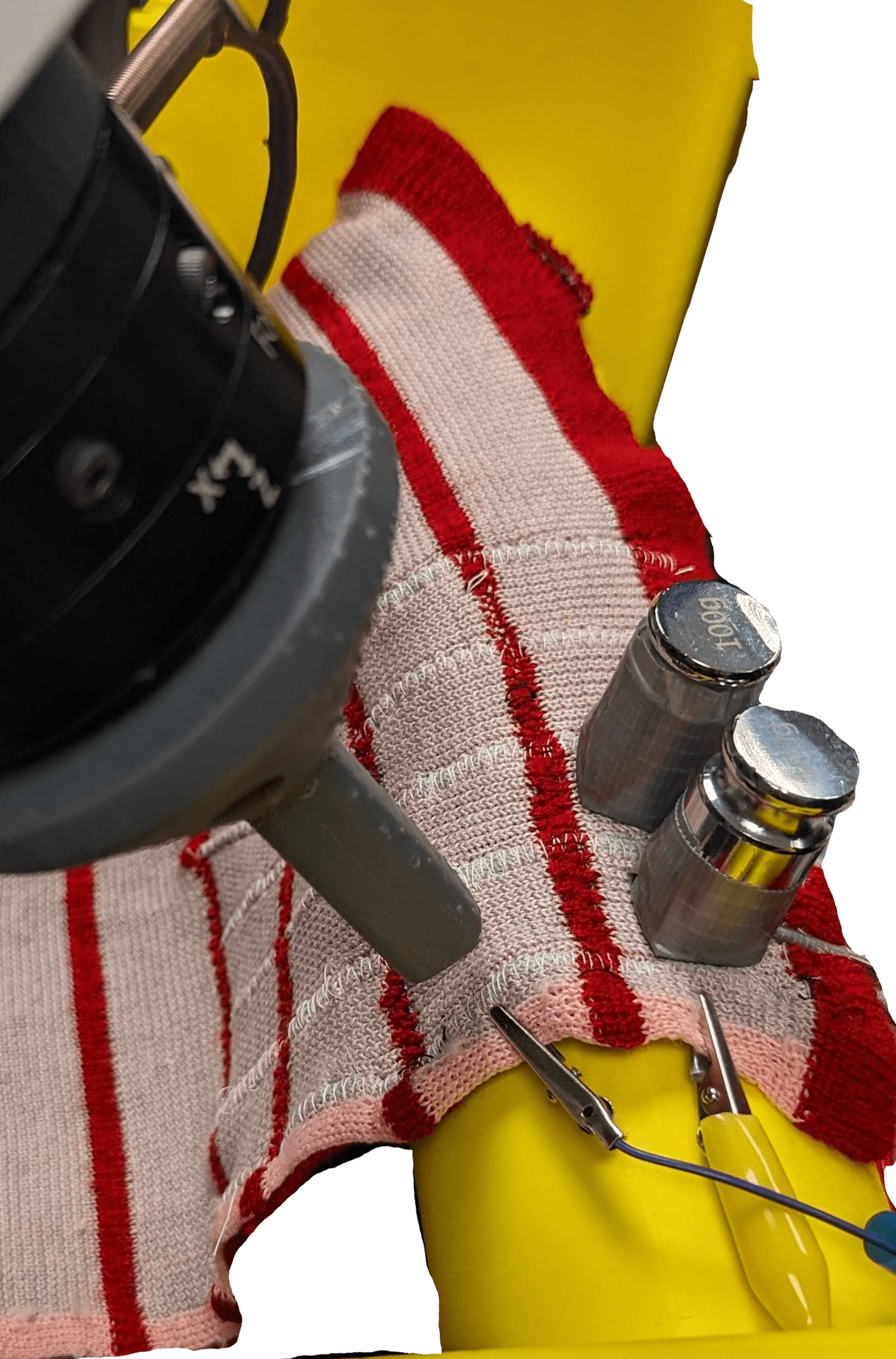}
        \caption{}
        
        \label{fig: curve-touch}
     \end{subfigure}
     \begin{subfigure}[t]{0.42\linewidth}
         \centering
        \includegraphics[width=\columnwidth]{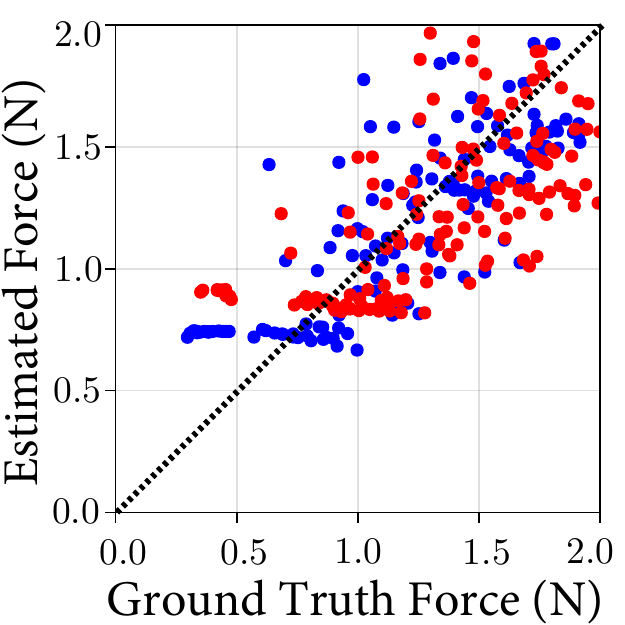}
        \caption{}
        
        \label{fig: curve-touch}
     \end{subfigure}
     \vspace{10pt}
     \caption{The Force Estimation experiment setup.  (a) flat surface experiment setup and (b)curved surface experiment. (c)  two-layer design of the skin allows the detection of tiny forces as low as 1N. Blue dots represent a single-point training set. Red dots represent a multi-touch test set.}
     \label{fig: setup-sensitivity}
\end{figure}

% Two 100g calibration weights are put on the skin the force torque sensor is pressed on another cell simultaneously. At any point in the sampling process, the ground truth force on all the cells is known, which is used to calculate RMSE.

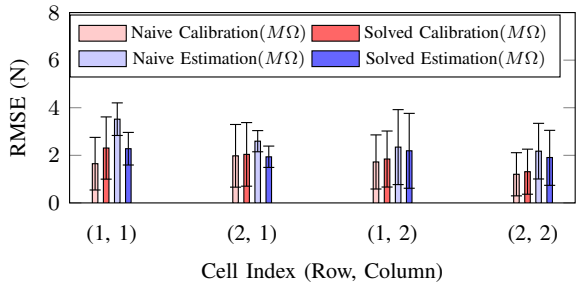
\begin{figure}[t!]
     \vspace{10pt}
        \centering
        \resizebox{0.9\linewidth}{!}{\input{Figures/flat_error_bar.tex}}

        \caption{Flat Surface Experiment. Single-touch force calibration is performed once to create a calibration set for the linear regression model. Our method achieved lower multi-touch error compared to the naive approach which predicts force from with raw voltage.}
        \label{fig:flat-error-bar}
\end{figure}%

\begin{figure}[t!]
        \centering
        \resizebox{0.9\linewidth}{!}{\input{Figures/curved_error_bar.tex}}

        \caption{Curved Surface Experiment. Single-touch force calibration is performed to create a calibration set for the linear regression model. Our method achieved lower multi-touch error compared to the naive approach which predicts force from with raw voltage.}
        \label{fig:curved-error-bar}
\end{figure}
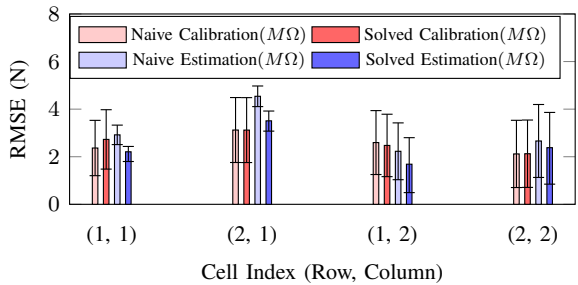%

\begin{figure}[t]
        % \begin{center}
        \centering
        \includegraphics[width=\columnwidth]{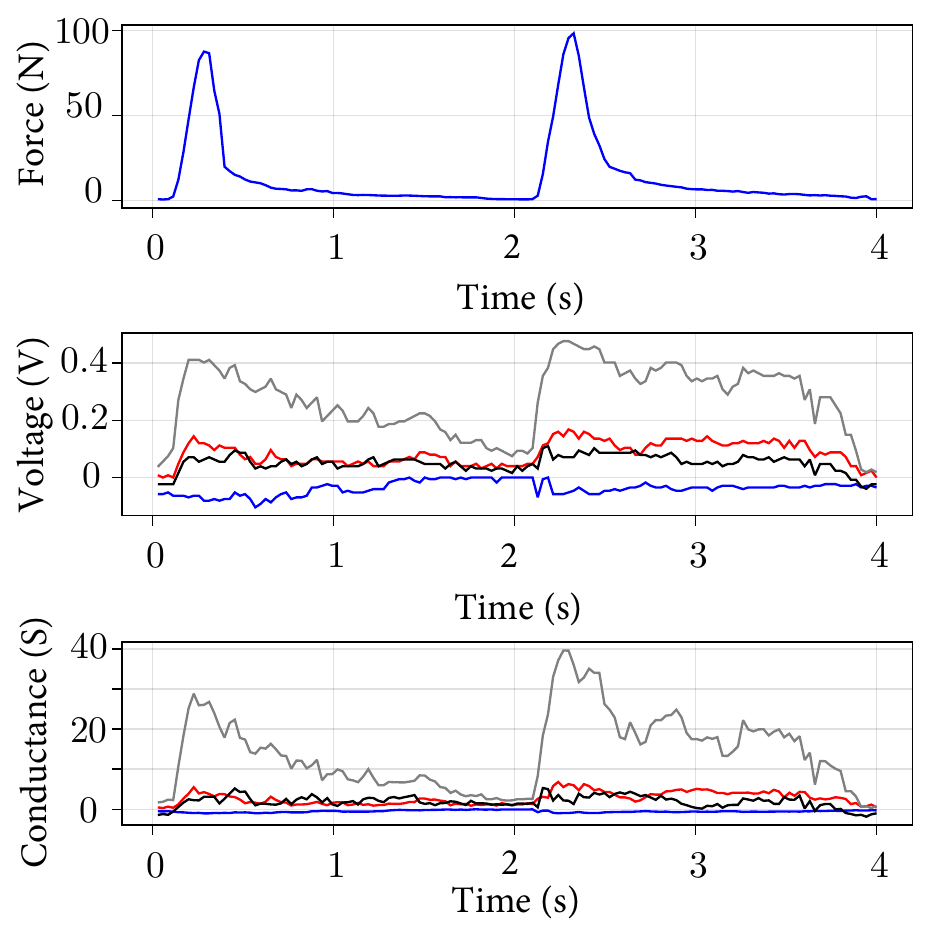}
        \caption{Time series data of 4 seconds with the robot tip pressing cell (2, 2). For the middle and bottom figures, Blue: cell (1, 1) and Red: cell (2, 1). Black: cell (1, 2). Gray: cell (2, 2). Top: Force Torque sensor readings at cells (2, 2). Middle: Raw voltage from the readout of the skin. Bottom: Solved conductance.}
        \vspace{-1\baselineskip}
        \label{fig: flat-line-solved}
        % \end{center}
\end{figure}

\begin{figure}

\end{figure}

\subsection{Force Estimation}

We conducted real-robot experiments using a Kinova Gen3 robot with a force-torque sensor to evaluate our method's force estimation accuracy, comparing it to ground truth measurements and an existing approach.

In the first experiment, the tactile skin was set on a flat surface and pressed by a robot's flat tip. The robot's tip moves gradually toward the skin, incurring varying levels of pressure. The pairs of data from both the force sensor and the tactile skin are used for calibration. After calibrating the sensor, single-point touches formed a calibration set through linear regression. Using this model, we estimated forces on cells (1, 2) upon placing two 100g weights on cells (1, 1) and (2, 1). With known ground truth forces, our method reduced the RMSE by 27.3\% compared to the existing ``Naive" approach, which directly estimates forces from voltages. The results are displayed in Figure \ref{fig:flat-error-bar}.

The skin was placed on a curved surface for the second experiment, altering its resistance distribution. Comparisons with the Naive approach revealed our method enhanced accuracy by 26.4\%, as seen in Figure~\ref{fig:curved-error-bar}.

\subsection{Time Series}
% noise suppression effects
% line plot 1 touch

We evaluated the tactile skin's time-series data using our algorithm. The force reading results shown in Figure \ref{fig: flat-line-solved} indicate that the algorithm effectively mitigates the ghosting effects caused by force applied on the same row or column, as evident from the reduced correlated noise in cells (1, 2) (Black) and (2, 1) (Red) due to the pressing cell (2, 2) (gray).

\subsection{Minimum Detectable Force}
Attributed to the two-layered design which removes the physical threshold presented by the middle layer, our proposed method could detect tiny force as low as 1N in the presence of multiple touches, shown in Figure \ref{fig: setup-sensitivity}.

\subsection{Scalability}
Our algorithm runs at 11 samples/second, on an AMD 7940HS Laptop CPU for a 2x2 skin, taking on average 30 iterations and 40ms for a least squares solution and 25 iterations and 50ms for the regularized solution. Textile skins as large as 16x16 have been manufactured and shown robust single-touch performance. However, on larger skins, such as 3x3 as shown in simulation, we are unable to achieve real-time performance due to the non-linear programming nature of the problem.
% Figure: Number of cells vs solving time

%%%%%%%%%%%%%%%%%%%%%%%%%%%%%%%%%%%%%%%%%%%%%%%%%%%%%%%%%%%%%%%%%%%%%%%%%%%%%%%%
\section{CONCLUSION}
% We formulate the problem as a circuit component parameter estimation problem. We solve an optimization problem with a Regularized least squares objective function for improved force sensing accuracy. This allows accurate readings even if the skin is dynamic (can be stretched, bent, etc.) Compared to the existing approach, we mitigate the ghosting effects and improved the estimate of the force values. Due to these improvements, we present a simplified skin design that facilitates easier manufacturing without compromising performance.
In this paper, we formulated multi-touch force sensing using textile-based tactile skins as a circuit parameter estimation problem. By employing a Regularized Least Squares objective function and optimization techniques, we mitigated the ghosting effects while achieving accurate multi-touch cell resistance estimations in both simulated and experimental environments. Our approach improved existing methods by reducing error rates in force prediction by 27.3\% on flat surfaces and 26.4\% on curved surfaces in multi-contact scenarios.

% Despite the fact that our approach takes wire resistances into consideration and is also robust to uneven resistance distributions of the tactile skin, we have observed that extensive noises in the voltage measurements can potentially hinder the accuracy of our circuit resistance estimation. The noise in voltage measurement can come from a variety of sources, e.g. the electric perturbation comes from a large AC power adapter. For the next step, we would like to improve the mechatronic design of our tactile skin and improve its robustness against 

%%%%%%%%%%%%%%%%%%%%%%%%%%%%%%%%%%%%%%%%%%%%%%%%%%%%%%%%%%%%%%%%%%%%%%%%%%%%%%%%
% \section*{APPENDIX}

% Appendixes should appear before the acknowledgment.

% \section*{ACKNOWLEDGMENT}

% The preferred spelling of the word acknowledgment in America is without an e after the g. Avoid the stilted expression, One of us (R. B. G.) thanks . . .  Instead, try R. B. G. thanks. Put sponsor acknowledgments in the unnumbered footnote on the first page.

%%%%%%%%%%%%%%%%%%%%%%%%%%%%%%%%%%%%%%%%%%%%%%%%%%%%%%%%%%%%%%%%%%%%%%%%%%%%%%%%
\bibliographystyle{IEEEtran}
\bibliography{root}
%%%%%%%%%%%%%%%%%%%%%%%%%%%%%%%%%%%%%%%%%%%%%%%%%%%%%%%%%%%%%%%%%%%%%%%%%%%%%%%%

\end{document}

%% file: Figures/simulation_wiresweep.tex
% \pgfplotstableread{bar_plot_data.dat}{\table}
\pgfplotstableread{
    0 0.7100171232727095 0.0 0.2210381299759276 0.0 0.08290191989016006 0.0
    1 0.7096245367535756 0.0 0.1911987672096791 0.0 0.06802354574122778 0.0
    2 0.7091340899694378 0.0 0.16138623530770824 0.0 0.05187655517268265 0.0
    3 0.6897616386413891 0.0 0.09798342293103046 0.0 0.11403515604174483 0.0
    4 0.6708240544005664 0.0 0.1893934230623146 0.0 0.19976386000520296 0.0
    
}\dataset
\begin{tikzpicture}
\begin{axis}[ybar, area legend,
        width=.99\textwidth,
        bar width=2pt,
        ymin=0,
        ymax=1.2,        
        ylabel={RMSE},
        xlabel={Wire Resistance $(M\Omega)$},
        xtick={0, 1, 2, 3, 4},
        xticklabels = {
            0.0001,
            0.0005,
            0.001,
            0.021,
            0.041
        },
        major x tick style = {opacity=0},
        % minor x tick num = 1,
        % minor tick length=2ex,
        legend style={at={(0.0,0.99)},anchor=north west,nodes={scale=0.8, transform shape}, font=\footnotesize},
        legend columns=2, font=\footnotesize, height = 4cm, width = 8cm
        ]
\addplot[draw=black,fill=red!20,error bars/.cd,y dir=both,y explicit] 
    table[x index=0,y index=1,y error plus index=2,y error minus index=2] \dataset; %Data1
    
\addplot[draw=black,fill=red!60,error bars/.cd,y dir=both,y explicit] 
    table[x index=0,y index=3,y error plus index=4,y error minus index=4] \dataset;
    
\addplot[draw=black,fill=blue!20,error bars/.cd,y dir=both,y explicit] 
    table[x index=0,y index=5,y error plus index=6,y error minus index=6] \dataset;
    
% \addplot[draw=black,fill=blue!60,error bars/.cd,y dir=both,y explicit] 
%     table[x index=0,y index=7,y error plus index=8,y error minus index=8] \dataset;

% \addplot[draw=black,fill=blue!60,error bars/.cd,y dir=both,y explicit] 
%     table[x index=0,y index=7,y error plus index=8,y error minus index=8] \dataset;

% \addplot[draw=black,fill=blue!60,error bars/.cd,y dir=both,y explicit] 
%     table[x index=0,y index=7,y error plus index=8,y error minus index=8] \dataset;
    
% \addplot[draw=black,fill=blue!60,error bars/.cd,y dir=both,y explicit] 
    % table[x index=0,y index=9,y error plus index=10,y error minus index=10] \dataset;
    %Data2
\legend{
    Naive Error $(M\Omega)$,
    % Wire Error Naive $(M\Omega)$,
    Least Squares Error $(M\Omega)$,
    % Wire Error Feasible $(M\Omega)$,
    Regularized Error $(M\Omega)$,
    % Wire Error Optimal $(M\Omega)$
    }
\end{axis}
\end{tikzpicture}

%% file: Figures/simulation_cellsweep.tex
% \pgfplotstableread{bar_plot_data.dat}{\table}
\pgfplotstableread{
0 0.7235412459927661 0.0 0.8071502873462489 0.0 0.8029565419067455 0.0
1 0.7127830317486055 0.0 0.4239527351484873 0.0 0.14888527872125093 0.0
2 0.7102470456074225 0.0 0.2127214638400493 0.0 0.06955408034103815 0.0
3 0.7156853194439154 0.0 0.018180476437598837 0.0 0.013814755425785969 0.0
    4 0.7878648518735506 0.0 0.03545990311485737 0.0 0.014501126802590096 0.0
}\dataset
\begin{tikzpicture}
\begin{axis}[ybar, area legend,
        width=.99\textwidth,
        bar width=2pt,
        ymin=0,     
        ymax=1.3,
        ylabel={RMSE $(M\Omega)$},
        xlabel={Cell Resistance $(M\Omega)$},
        xtick={0, 1, 2, 3, 4},
        xticklabels = {
            0.01,
            0.07,
            0.09,
            0.5,
            0.7
        },
        major x tick style = {opacity=0},
        % minor x tick num = 1,
        % minor tick length=2ex,
        legend style={at={(0.0,0.99)},anchor=north west,nodes={scale=0.8, transform shape}, font=\footnotesize},
        legend columns=2, font=\footnotesize, height = 4cm, width = 8cm
        ]
\addplot[draw=black,fill=red!20,error bars/.cd,y dir=both,y explicit] 
    table[x index=0,y index=1,y error plus index=2,y error minus index=2] \dataset; %Data1
    
\addplot[draw=black,fill=red!60,error bars/.cd,y dir=both,y explicit] 
    table[x index=0,y index=3,y error plus index=4,y error minus index=4] \dataset;
    
\addplot[draw=black,fill=blue!20,error bars/.cd,y dir=both,y explicit] 
    table[x index=0,y index=5,y error plus index=6,y error minus index=6] \dataset;
    
% \addplot[draw=black,fill=blue!60,error bars/.cd,y dir=both,y explicit] 
%     table[x index=0,y index=7,y error plus index=8,y error minus index=8] \dataset;

% \addplot[draw=black,fill=blue!60,error bars/.cd,y dir=both,y explicit] 
%     table[x index=0,y index=7,y error plus index=8,y error minus index=8] \dataset;

% \addplot[draw=black,fill=blue!60,error bars/.cd,y dir=both,y explicit] 
%     table[x index=0,y index=7,y error plus index=8,y error minus index=8] \dataset;
    
% \addplot[draw=black,fill=blue!60,error bars/.cd,y dir=both,y explicit] 
    % table[x index=0,y index=9,y error plus index=10,y error minus index=10] \dataset;
    %Data2
\legend{
    Naive Error $(M\Omega)$,
    % Wire Error Naive $(M\Omega)$,
    Least Squares Error $(M\Omega)$,
    % Wire Error Feasible $(M\Omega)$,
    Regularized Error $(M\Omega)$,
    % Wire Error Optimal $(M\Omega)$
    }
\end{axis}
\end{tikzpicture}

%% file: Figures/flat_error_bar.tex
% 2.7977660624256533 / 1.9562295015649027
% 2.967742406151791 / 1.7411714066305304
% 4.752793614738147 / 4.458493686408567
% 4.271169515247027 / 3.7313344333787786
% \pgfplotstableread{bar_plot_data.dat}{\table}
\pgfplotstableread{
    % 0 1.1850477352973459 0.0014953286702253962 1.318199891890796 0.0016633440432150494 3.1345229775732815 0.00020418664163263317 2.941718094586462 0.00012058209928794452
    % 1 1.012772326333349 0.002073230003772161 1.0468281180646786 0.002142945069422927 5.24198554054568 0.00023450807597665075 5.0804036626502995 0.0001712753047364087
    % 2 0.8231612369592937 0.002375648178861984 0.7769027649557364 0.0022421459561648488 0.7436596267409239 0.0018225577163972447 0.6915749329212457 0.0017003443903495956
    % 3 0.7103679725264996 0.002348327084000843 0.679317438456406 0.0022456805501626753 0.880772180091415 0.0025927583143166293 0.8639040236217629 0.0024496807541189965

    % 0 3.3165619564674516 0.005495547795747396 3.335501786302142 0.005526931120246096 2.7977660624256533 0.0010465036715620105 1.9562295015649027 0.0010084147752153416
    % 1 4.788545962136928 0.006711348476407025 4.502442001878702 0.00631036174829464 2.967742406151791 0.0010080077720290585 1.7411714066305304 0.0009622332538837335
    % 2 3.956133174389202 0.004923626559246828 3.5859808168913525 0.004462951476278819 4.752793614738147 0.00862434028767787 4.458493686408567 0.008199934155048974
    % 3 4.9194299256671385 0.007839731453971122 4.885700780834247 0.007785979811675677 4.271169515247027 0.023007928072803545 3.7313344333787786 0.021898709715868547
    
0	1.648273502	1.105475197	2.306563111	1.307726834	3.519282977	0.6849066772	2.278219083	0.6845120676
1	1.979307437	1.316941135	2.039719114	1.33688767	2.594386924	0.4448001021	1.936295337	0.45202231
2	1.721372328	1.137123655	1.844507648	1.177092291	2.345398157	1.575148917	2.191151962	1.571585963
3	1.203957583	0.906926764	1.312178924	0.9468106929	2.175024038	1.169019723	1.907833698	1.142053509

}\dataset
\begin{tikzpicture}
\begin{axis}[ybar, area legend,
        width=.99\textwidth,
        bar width=2pt,
        ymin=0,
        ymax=8,        
        ylabel={RMSE (N)},
        xlabel={Cell Index (Row, Column)},
        xtick={0, 1, 2, 3},
        xticklabels = {
            (1, 1),
            (2, 1),
            (1, 2),
            (2, 2),
        },
        major x tick style = {opacity=0},
        % minor x tick num = 1,
        % minor tick length=2ex,
        legend style={at={(0.0,0.99)},anchor=north west,nodes={scale=0.8, transform shape}, font=\footnotesize},
        legend columns=2, font=\footnotesize, height = 4cm, width = 8cm
        ]
\addplot[draw=black,fill=red!20,error bars/.cd,y dir=both,y explicit] 
    table[x index=0,y index=1,y error plus index=2,y error minus index=2] \dataset; %Data1
    
\addplot[draw=black,fill=red!60,error bars/.cd,y dir=both,y explicit] 
    table[x index=0,y index=3,y error plus index=4,y error minus index=4] \dataset;
    
\addplot[draw=black,fill=blue!20,error bars/.cd,y dir=both,y explicit] 
    table[x index=0,y index=5,y error plus index=6,y error minus index=6] \dataset;
    
\addplot[draw=black,fill=blue!60,error bars/.cd,y dir=both,y explicit] 
    table[x index=0,y index=7,y error plus index=8,y error minus index=6] \dataset;
    
% \addplot[draw=black,fill=blue!60,error bars/.cd,y dir=both,y explicit] 
%     table[x index=0,y index=7,y error plus index=8,y error minus index=8] \dataset;

% \addplot[draw=black,fill=blue!60,error bars/.cd,y dir=both,y explicit] 
%     table[x index=0,y index=7,y error plus index=8,y error minus index=8] \dataset;

% \addplot[draw=black,fill=blue!60,error bars/.cd,y dir=both,y explicit] 
%     table[x index=0,y index=7,y error plus index=8,y error minus index=8] \dataset;
    
% \addplot[draw=black,fill=blue!60,error bars/.cd,y dir=both,y explicit] 
    % table[x index=0,y index=9,y error plus index=10,y error minus index=10] \dataset;
    %Data2
\legend{
    Naive Calibration$(M\Omega)$,
    Solved Calibration$(M\Omega)$,
    Naive Estimation$(M\Omega)$,
    Solved Estimation$(M\Omega)$,
    }
\end{axis}
\end{tikzpicture}

%% file: Figures/curved_error_bar.tex
% 2.7977660624256533 / 1.9562295015649027
% 2.967742406151791 / 1.7411714066305304
% 4.752793614738147 / 4.458493686408567
% 6.820893387373494 / 6.59922845657676
% \pgfplotstableread{bar_plot_data.dat}{\table}
\pgfplotstableread{
    % 0 2.3661496282615766 0.003377801900085695 2.728099052484457 0.0038945035652180123 2.9194008546711667 0.00041832175055892185 2.2080297584604143 0.00012541950606128518
    % 1 3.1223553149609735 0.004656757916646294 3.119595262059236 0.004652641505506619 4.540025091803753 0.0004689686858785983 3.5087752902242437 0.00042314310306820795
    % 2 2.5954408305094185 0.004509890199945657 2.474711516731456 0.004300108515596286 2.228367495000596 0.003562956726074259 1.685945950295575 0.003108044861451178
    % 3 2.1181022474750812 0.00498964362013283 2.1266100203552187 0.005009685501833073 2.6621659939051696 0.005875054573545554 2.380487326858459 0.005485788431364103

0	2.366149628	1.162377202	2.728099052	1.248119155	2.919400855	0.4090583091	2.208029758	0.2239817011
1	3.122355315	1.364808839	3.119595262	1.364205484	4.540025092	0.4331136968	3.50877529	0.4114088492
2	2.595440831	1.343114321	2.474711517	1.311504253	2.228367495	1.193810157	1.68594595	1.114996836
3	2.118102247	1.41274819	2.12661002	1.415582636	2.662165994	1.532978092	2.380487327	1.48132217
}\dataset
\begin{tikzpicture}
\begin{axis}[ybar, area legend,
        width=.99\textwidth,
        bar width=2pt,
        ymin=0,
        ymax=8,        
        ylabel={RMSE (N)},
        xlabel={Cell Index (Row, Column)},
        xtick={0, 1, 2, 3},
        xticklabels = {
            (1, 1),
            (2, 1),
            (1, 2),
            (2, 2),
        },
        major x tick style = {opacity=0},
        % minor x tick num = 1,
        % minor tick length=2ex,
        legend style={at={(0.0,0.99)},anchor=north west,nodes={scale=0.8, transform shape}, font=\footnotesize},
        legend columns=2, font=\footnotesize, height = 4cm, width = 8cm
        ]
\addplot[draw=black,fill=red!20,error bars/.cd,y dir=both,y explicit] 
    table[x index=0,y index=1,y error plus index=2,y error minus index=2] \dataset; %Data1
    
\addplot[draw=black,fill=red!60,error bars/.cd,y dir=both,y explicit] 
    table[x index=0,y index=3,y error plus index=4,y error minus index=4] \dataset;
    
\addplot[draw=black,fill=blue!20,error bars/.cd,y dir=both,y explicit] 
    table[x index=0,y index=5,y error plus index=6,y error minus index=6] \dataset;
    
\addplot[draw=black,fill=blue!60,error bars/.cd,y dir=both,y explicit] 
    table[x index=0,y index=7,y error plus index=8,y error minus index=6] \dataset;
    
% \addplot[draw=black,fill=blue!60,error bars/.cd,y dir=both,y explicit] 
%     table[x index=0,y index=7,y error plus index=8,y error minus index=8] \dataset;

% \addplot[draw=black,fill=blue!60,error bars/.cd,y dir=both,y explicit] 
%     table[x index=0,y index=7,y error plus index=8,y error minus index=8] \dataset;

% \addplot[draw=black,fill=blue!60,error bars/.cd,y dir=both,y explicit] 
%     table[x index=0,y index=7,y error plus index=8,y error minus index=8] \dataset;
    
% \addplot[draw=black,fill=blue!60,error bars/.cd,y dir=both,y explicit] 
    % table[x index=0,y index=9,y error plus index=10,y error minus index=10] \dataset;
    %Data2
\legend{
    Naive Calibration$(M\Omega)$,
    Solved Calibration$(M\Omega)$,
    Naive Estimation$(M\Omega)$,
    Solved Estimation$(M\Omega)$,
    }
\end{axis}
\end{tikzpicture}